%% file: main.tex
\documentclass[10pt,twocolumn,letterpaper]{article}

\usepackage[pagenumbers]{iccv} 
\usepackage[accsupp]{axessibility}

\usepackage{algorithm}
\usepackage{algorithmic}
\usepackage{graphicx, subcaption}
\usepackage{multirow}
\usepackage{array}
\usepackage{makecell} 

\input{preamble}

\definecolor{iccvblue}{rgb}{0.21,0.49,0.74}
\usepackage[pagebackref,breaklinks,colorlinks,allcolors=iccvblue]{hyperref}

\def\confName{ICCV}
\def\confYear{2025}

\title{\textit{DoppDrive}: Doppler-Driven Temporal Aggregation \\ for Improved Radar Object Detection}

\author{Yuval Haitman\footnotemark[1] \space  and Oded Bialer\thanks{Both authors contributed equally to this work.\\ Both authors are with General Motors, Yuval Haitman is also with the School of Electrical and Computer Engineering in Ben Gurion University of the Negev.} \\
General Motors, 
Technical Center Israel\\
{\tt\small haitman@post.bgu.ac.il, \tt\small oded.bialer8@gmail.com}}

\begin{document}
\maketitle
\input{0_abstract}    
\input{1_intro}

\input{2_related_work}
\input{3_method}

\input{4_results}
\input{5_conclusion}
{
    \small
    \bibliographystyle{ieeenat_fullname}
    \bibliography{main}
}

\newpage
\appendix
\maketitlesupplementary
\input{6_supp}

\end{document}

%% file: preamble.tex
\usepackage{pifont} 

\newcommand{\V}{\ding{51}} 
\newcommand{\X}{\ding{55}} 

%% file: 0_abstract.tex
\begin{abstract}
Radar-based object detection is essential for autonomous driving due to radar's long detection range. However, the sparsity of radar point clouds, especially at long range, poses challenges for accurate detection. Existing methods increase point density through temporal aggregation with ego-motion compensation, but this approach introduces scatter from dynamic objects, degrading detection performance. We propose \textit{DoppDrive}, a novel Doppler-Driven temporal aggregation method that enhances radar point cloud density while minimizing scatter. Points from previous frames are shifted radially according to their dynamic Doppler component to eliminate radial scatter, with each point assigned a unique aggregation duration based on its Doppler and angle to minimize tangential scatter. \textit{DoppDrive} is a point cloud density enhancement step applied before detection, compatible with any detector, and we demonstrate that it significantly improves object detection performance across various detectors and datasets.Our project page: \href{https://yuvalhg.github.io/DoppDrive/}{https://yuvalhg.github.io/DoppDrive/}
\end{abstract}

%% file: 1_intro.tex
\section{Introduction}
\label{sec:intro}
Radar plays an important role in autonomous driving and active safety systems due to its capability to provide detection at long range \cite{ignatious2022overview,yurtsever2020survey,marti2019review}. While some approaches use neural networks to detect objects directly from the radar's reflection intensity tensor spectrum \cite{zhang2021raddet,zhang2020object,dong2020probabilistic,kim2020yolo,haitman2024boostrad}, spanning spatial and Doppler domains, this low-level data is not always accessible. When available, it requires substantial communication bandwidth, memory, and computational resources to process. A more efficient alternative is object detection from radar point cloud data \cite{palffy2022multi,liu2023spatial,popov2023nvradarnet,paek2023enhanced,rebut2022raw,fent2023radargnn,tan20223}, where the point cloud is generated by filtering high-probability reflection points from the intensity tensor spectrum \cite{rohling1983radar}. This paper focuses on object detection using radar point clouds, a method conceptually similar to LIDAR-based detection but with critical distinctions.
\begin{figure}[t]
\centering
\includegraphics[width=1\columnwidth]{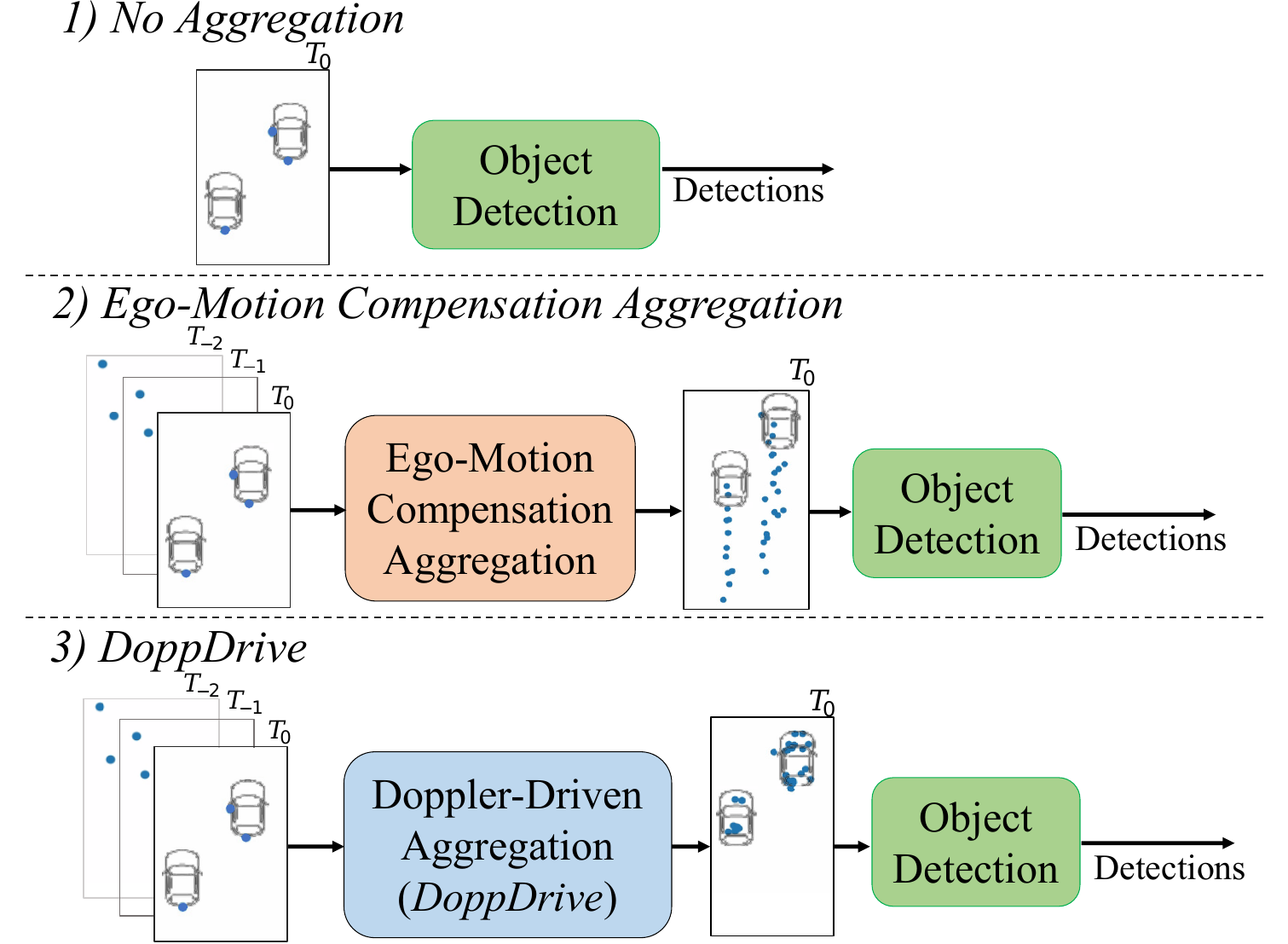}
\caption{\textbf{Object detection with different input point clouds.} (1) single frame, highly sparse point cloud, (2) temporally aggregated with ego-motion compensation—scattered points from dynamic objects, (3) Doppler-driven aggregation—dense points with minimal scatter, enhancing object detection.}
\label{fig:teaser}
\end{figure}

\begin{figure*}[h]
\centering
\includegraphics[width=0.8\textwidth]{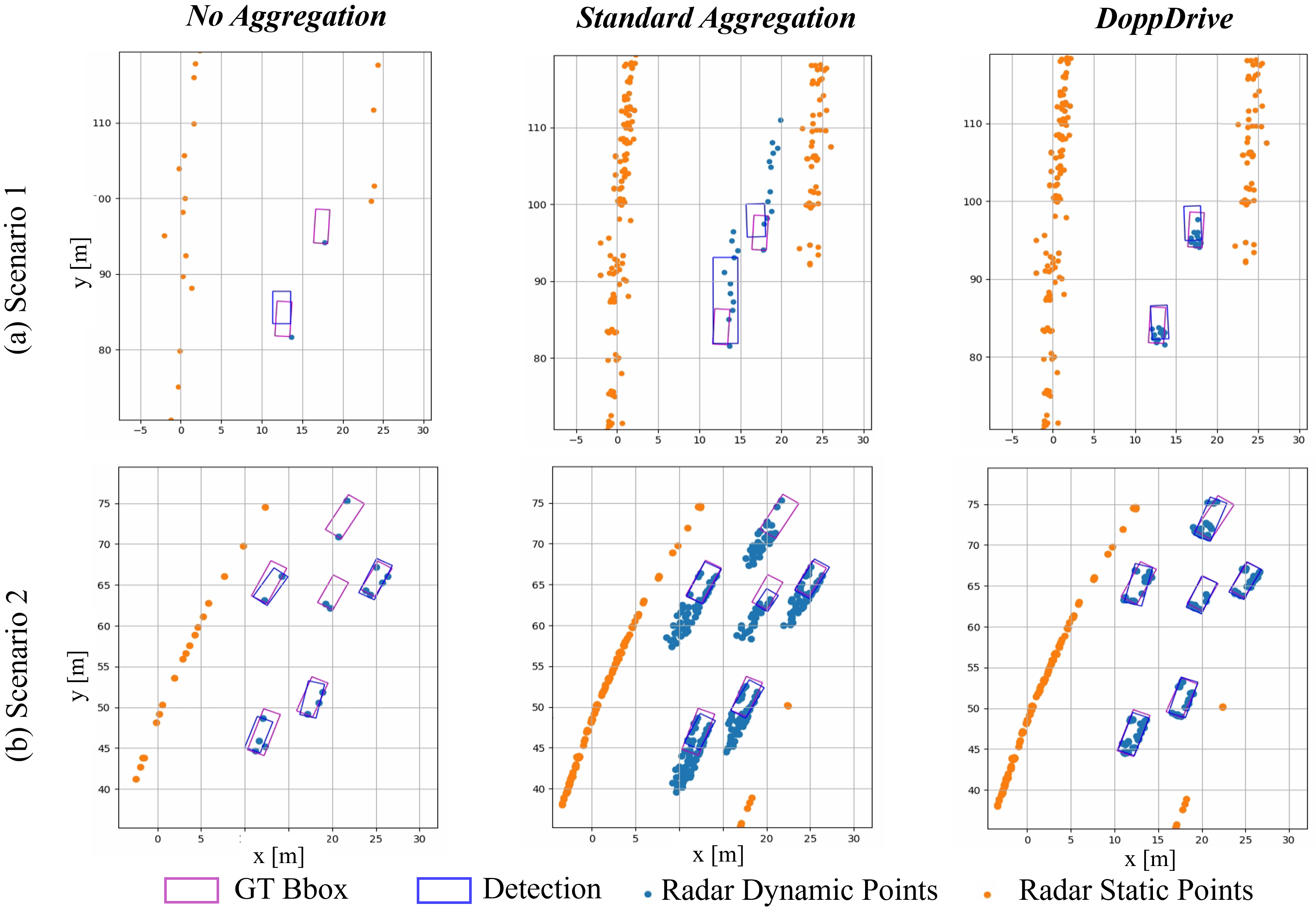}
\caption{
\textbf{Qualitative examples demonstrating \textit{DoppDrive} in two scenarios:} the upper row (a) from the aiMotive dataset \cite{matuszka2022aimotive} and the lower row (b) from \textit{LRR-Sim} (\cref{sec:LLR_SIM}). Columns show: (left) Sparse point cloud without aggregation, leading to missed detections; (center) Aggregation with ego-motion compensation, causing scattered points around dynamic objects and inaccuracies in object size and position; (right) \textit{DoppDrive} enhances density around dynamic objects (blue points) with minimal scatter, improving detection accuracy. Detections are obtained using SMF \cite{liu2023spatial}.}
\label{fig:qual_example}
\vspace{-10pt}
\end{figure*}

One key difference between radar and LIDAR point clouds is radar’s much lower density, due to its lower angular resolution and longer wavelength \cite{giuffrida2023survey}. This sparsity increases with distance, resulting in only a few reflections from distant objects, making it challenging to distinguish true objects from noise and accurately detect their shape, orientation, and size. 
As objects' reflection points vary with changes in the relative position between objects and the radar, a common approach to increasing point cloud density and improving object detection is to aggregate point clouds from multiple frames over a short duration (about 0.5 seconds), applying ego-motion compensation and adding a timestamp to each point \cite{palffy2022multi, tan20223, sless2019road, popov2023nvradarnet, liu2023spatial, yilmaz2024mask4former}. While this method aligns reflections from static objects, the unknown velocities of dynamic objects cause significant scatter in the aggregated data, limiting performance. 

Developing methods to increase radar point cloud density without introducing scatter could greatly improve object detection, especially for long-range radar, which must detect objects at distances of up to $300m$ for high-speed autonomous driving \cite{continental_ars540,bosch_front_radar,zf_4d_radar_2021,aptiv_radars,sun2020mimo,markel2022radar,zhou2020mmw}. However, current automotive radar datasets are limited to $175m$, primarily due to the difficulty of obtaining accurate annotations at greater distances. At these ranges, radar points become very sparse, and the LIDAR and camera sensors typically used for annotation struggle to detect objects accurately. A dataset extending beyond $175m$ with precise annotations could significantly advance research in long-range radar detection.

Another key distinction between radar and LIDAR is radar’s Doppler measurement, which captures the radial velocity of each reflection point relative to the radar. 
The Doppler measurement consists of two components: the \textit{ego-speed Doppler} component (due to ego-vehicle motion) and the \textit{dynamic Doppler} component (due to object motion). The ego-speed component can be calculated from the ego-vehicle’s velocity \cite{tan20223}, which is either obtained from auxiliary sensors like GPS-INS or odometers \cite{farrell1999global,alaba2024gps,spilker1996global,chatfield1997fundamentals}
or can be estimated from the radar's point cloud Doppler information \cite{tan20223,kellner2013instantaneous,almalioglu2020milli,rapp2015fast}. Previous work on radar-based object detection incorporates Doppler information as an additional input feature for each point, using object detection networks originally designed for LIDAR point clouds with the added Doppler feature \cite{palffy2022multi,liu2023spatial,popov2023nvradarnet,tan20223}. 
Removing the ego-speed Doppler component from the Doppler measurement has been shown to aid object detection \cite{palffy2022multi,tan20223}, as it enables the residual Doppler to indicate whether a point reflects a static (zero residual Doppler) or dynamic (non-zero residual Doppler) object, helping the neural network learn distinct features for detection. However, exploring additional ways to leverage the dynamic Doppler component information to enhance object detection yet remains a valuable and important direction for research.

This paper introduces \textit{DoppDrive}, a Doppler-driven temporal aggregation method that uses Doppler information in a novel way to enhance radar point cloud prior to object detection. \textit{DoppDrive} increases point cloud density while minimizing scatter in reflections from dynamic objects. By aligning points from previous frames to the current frame along the radial direction based on dynamic Doppler velocity, \textit{DoppDrive} eliminates range scatter. Additionally, it mitigates tangential spread by setting each point’s aggregation duration based on its specific Doppler and angle, resulting in dynamic aggregation durations per point rather than a fixed duration for all. 

\cref{fig:teaser} shows the high-level concept of \textit{DoppDrive} compared to standard aggregation with ego-vehicle motion compensation and no aggregation. \cref{fig:qual_example} presents qualitative examples of these point clouds, demonstrating \textit{DoppDrive}'s increased density with minimal dispersion. As we will show, this enhancement improves object detection performance. \textit{DoppDrive} is compatible with any point cloud detection network, and we validate its performance improvements across four detectors and three datasets.

To address the shortage of datasets for testing methods on sparse radar point clouds beyond $200m$, we introduce a new dataset, \textit{LRR-Sim}, as part of this work. \textit{LRR-Sim} simulates a long-range automotive radar in highway scenarios, providing annotated point clouds up to $300m$. We use it to evaluate our method alongside reference approaches at extended detection ranges. The dataset is publicly released to support future research in long-range radar perception.

The main contributions of the paper are as follows:
\begin{enumerate}
\item A novel Doppler-driven temporal aggregation method that increases the density of radar point clouds while minimizing spatial scattering of points from dynamic objects, thereby enhancing object detection performance.

\item The release of \textit{LRR-Sim}, a long-range automotive radar dataset with sparse point clouds from extended detection ranges and precise annotations up to $300m$. It is generated by radar simulation to address the challenge of obtaining accurate long-distance annotations. 
The dataset is publicly available and serves as a valuable resource for advancing radar-based long-range detection.
\end{enumerate}

%% file: 2_related_work.tex
\section{Related Work}
\label{sec:related_work}

\subsection{Object Detection From Radar Point Cloud}
Various methods exist for radar point cloud object detection. Several methods leverage PointPillars \cite{lang2019pointpillars} for radar detection \cite{zheng2023rcfusion,palffy2022multi,liu2023spatial,tan20223,xu2021rpfa}, while NVRadarNet \cite{popov2023nvradarnet} uses 2D convolution with a Feature Pyramid Network \cite{lin2017feature} in bird’s-eye view, and K-Radar \cite{paek2023enhanced} applies 3D sparse convolution. RADIal \cite{rebut2022raw} utilizes PIXOR \cite{yang2018pixor}, and Radar-transformer \cite{bai2021radar} employs a transformer architecture for radar detection. Additional approaches include graph convolution \cite{svenningsson2021radar,fent2023radargnn} and hybrid of point-based and grid-based networks \cite{ulrich2022improved}.
All radar point cloud detection networks face challenges due to point cloud sparsity, which limits performance. 
Increasing point cloud density can enhance their effectiveness, which is the focus of our work.

\subsection{Radar Datasets}\label{sec:radar_datasets}
Radar point cloud datasets are characterized by detection range and angular resolution, where higher resolution results in a denser point cloud, while a longer detection range reduces point density.

Publicly available low-resolution radar datasets with object annotations include nuScenes \cite{caesar2020nuscenes}, covering urban, residential, and industrial areas; PixSet \cite{deziel2021pixset}, focused on high-density urban environments, both limited to object ranges below $80m$; and aiMotive \cite{matuszka2022aimotive}, which extends detection up to $175m$ in urban and highway environments. High-resolution radar datasets with object annotations include K-Radar \cite{paek2023enhanced}, which has limited Doppler speed measurements; View-of-Delft \cite{palffy2022multi}, with a maximum detection range of $50m$; Astyx \cite{meyer2019automotive}, which is relatively small; TJ4DRadSet \cite{zheng2022tj4dradset}, with a detection range up to $90m$; and RADIal \cite{rebut2022raw}, which has a $103m$ range and captures 
data from highway, country-side and city driving. 

In this paper, we evaluate our method performance on the aiMotive and RADIal datasets, representing low- and high-resolution radars with relatively long detection ranges. Additionally, we introduce a new dataset for long-range radar detection, with point clouds and objects up to $300m$, significantly extending the range of existing datasets (see \cref{sec:LLR_SIM}).

\subsection{Ego-Vehicle Velocity}\label{sec:radar_ego_vel_est}
Ego-vehicle velocity is typically obtained from a GPS-INS sensor or vehicle odometers \cite{farrell1999global,alaba2024gps,spilker1996global,chatfield1997fundamentals}. Alternatively, it can be estimated from radar point clouds by identifying static objects (e.g., guardrails, buildings, poles, ground) via Doppler clustering. The Doppler of each static point represents a different radial component of the vehicle's velocity. Thus, Doppler measurements from static reflections can be expressed as a function of point angles and ego-vehicle velocity, allowing the velocity to be resolved \cite{tan20223,kellner2013instantaneous,almalioglu2020milli,rapp2015fast}.

%% file: 3_method.tex
\section{Background: Doppler Components}\label{sec:dopp_background}
\begin{figure}[t]
\centering
\includegraphics[width=0.2\textwidth]{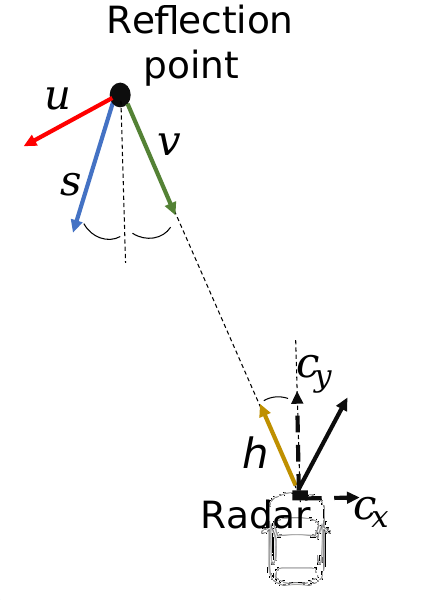}
\caption{\textbf{Scenario with a dynamic reflection point and a moving radar.} The reflection point’s radial and tangential velocity components are \( v \) and \( u \), respectively. The Doppler measurement is the sum of the radial velocity \( v \) of the dynamic object (dynamic Doppler component) and the radial velocity due to ego-vehicle motion \( h \) (ego-speed Doppler component).}
\label{fig:dopp_components}
\end{figure}

This section provides background material used in deriving our method in  \cref{sec:method}.
In automotive scenarios, the variation in height over short aggregation durations is relatively small. Thus, throughout this paper, we assume zero vertical velocity and focus only on longitudinal and lateral velocity.
\cref{fig:dopp_components} shows a reflection point in top view relative to the radar, where the point’s angle to the radar is \( \theta \) and its heading angle is \( \alpha \) with speed \( s \) (blue arrow). The radial and tangential velocity components are \( v \) and \( u \) (green and red arrows), respectively, and are given by:
\begin{equation}\label{eq:rad_vel}
v = s \cos(\theta + \alpha),
\end{equation}
\begin{equation}\label{eq:tan_vel}
u = s \sin(\theta + \alpha).
\end{equation}
Solving for \( s \) in \cref{eq:rad_vel} and substituting into \cref{eq:tan_vel}, we obtain:
\begin{equation}\label{eq:rad_tan_components}
u = v \tan(\theta + \alpha).
\end{equation}

In \cref{fig:dopp_components}, the radar-equipped host vehicle moves along the black arrow with speed vector $\boldsymbol{c}=[c_x,c_y]$. 
The projection of $\boldsymbol{c}$ onto the reflection point’s direction is denoted by $h$, and calculated by:
\begin{equation}\label{eq:ego_speed_dopp}
h = c_x \sin(\theta) + c_y \cos(\theta).
\end{equation}
The radar measures Doppler frequency, which is converted to velocity units.
The point's Doppler measurement, denoted by \( d \), is the relative radial velocity between the radar and the point, accounting for both the point and ego-vehicle motions: \( d = v + h \), where \( v \) is the dynamic Doppler component (point’s radial velocity) and \( h \) is the ego-speed Doppler component.

The speed of the ego vehicle $\boldsymbol{c}$ can be obtained from a GPS-INS sensor or estimated from the radar point cloud (see \cref{sec:radar_ego_vel_est}). 
Given $\boldsymbol{c}$, we calculate $h$ for each point from \cref{eq:ego_speed_dopp} and obtain its dynamic Doppler component as $v=d-h$.
In \cref{sec:method}, we represent the reflection point’s velocity vector using the radial component \( v \) (dynamic Doppler component) and the tangential component from \cref{eq:rad_tan_components}.

\section{Doppler-Driven Temporal Aggregation}\label{sec:method}
Our point cloud aggregation algorithm enhances reflection point density to improve object detection by combining radar point clouds from previous frames with those of the current frame, as shown in \cref{fig:teaser}. This aggregation adjusts for both the ego-vehicle’s motion and the movement of dynamic objects between frames, using Doppler information from the radar point clouds.

\begin{figure}[t]
\centering
\includegraphics[width=\linewidth]{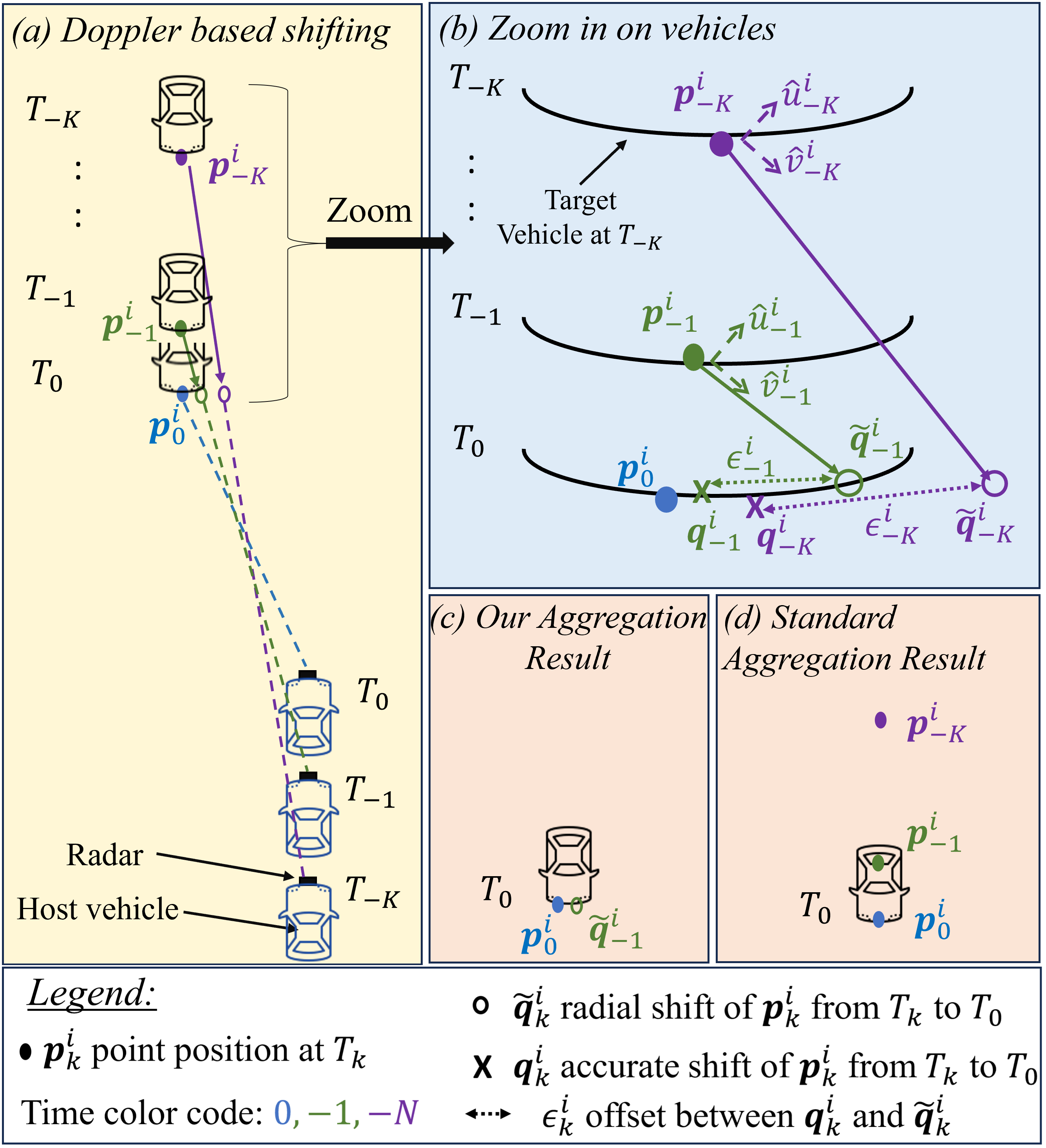}
    \vspace{-15pt}
\caption{\textbf{Overview of the Doppler-driven temporal aggregation method.} (a) Top view of the scenario with the radar-equipped host vehicle and a dynamic vehicle moving toward each other. Points from previous frames \( T_{-1} \) and \( T_{-K} \) are aggregated with the current frame \( T_0 \), compensating for the dynamic vehicle's motion by shifting points along the radial direction based on their dynamic Doppler measurement. (b) Zoom-in on (a), showing the accurate shift \( \boldsymbol{q}^i_k \) vs. the approximated shift \( \tilde{\boldsymbol{q}}^i_k \), with resulting offset distance \( \epsilon^i_k \). (c) Result of our aggregation method, showing denser reflection points with minimal disparity where the aggregation duration of \( \tilde{\boldsymbol{q}}^i_{-K} \) yields an expected offset beyond the tolerated limit, so it is discarded. (d) Result of standard aggregation, showing significant scatter in the aggregated points.}
\label{fig:agg_method}
\end{figure}

\cref{fig:agg_method} illustrates the proposed Doppler-Driven aggregation method. In the bird's-eye-view scenario in \cref{fig:agg_method}(a), the radar-equipped host vehicle (bottom) and an oncoming dynamic vehicle approach each other. Frame times are denoted by \( T_k \), where \( k=0 \) is the current frame and negative \( k \) values indicate earlier frames. The figure shows both vehicles at three frames: the current frame \( T_0 \), one prior frame \( T_{-1} \), and $K$ frames back at \( T_{-K} \). Let \( \boldsymbol{p}^{i}_{k} \) represent the 3D position $(x, y, z)$ of the \( i \)-th reflection point at frame \( T_k \). In this example, a single reflection point from the approaching vehicle appears in each frame, represented as \( \boldsymbol{p}^{i}_{-K} \), \( \boldsymbol{p}^{i}_{-1} \), and \( \boldsymbol{p}^{i}_{0} \), shown in top view as solid circles in purple, green, and blue, respectively, in \cref{fig:agg_method}.
Throughout this section, all point positions are referenced to a \emph{unified coordinate system}, such as the radar's coordinate system at $T_0$. This is achieved by transforming all points to $T_0$ coordinates, using ego-vehicle motion data from external sensors or estimating it from the radar point cloud (see \cref{sec:radar_ego_vel_est}).

The radial direction between each reflection point, $\boldsymbol{p}^{i}_{k}$, and the radar position at frame time $T_k$ is shown as dashed lines in \cref{fig:agg_method}(a). Our key idea is to compensate for the shift in reflection points from previous frames to the current frame caused by the dynamic vehicle’s motion, using the dynamic Doppler measurement, which represents the dynamic vehicle's radial velocity. Each previous point is shifted along its radial direction according to its dynamic Doppler measurement, as illustrated by solid arrows in \cref{fig:agg_method}(a), with arrow colors matching the points. The adjusted positions, marked by hollow circles in corresponding colors, correct for radial motion but still leave a tangential offset. To mitigate this, we limit each point’s aggregation duration, as explained further below.

To further explain our aggregation method, we refer to \cref{fig:agg_method}(b), which provides a zoomed-in view of \cref{fig:agg_method}(a). Let \( \boldsymbol{q}^{i}_k \) be the position of \( \boldsymbol{p}^{i}_k \) after shifting it accurately to the current time frame \( T_0 \) based on the dynamic object's precise velocity. These shifted points are marked with an ‘x’ in \cref{fig:agg_method}(b), with colors matching their original positions. 
We neglect any slight motion in height during the short integration interval and approximate zero velocity in this dimension.
Thus, the velocity vector that accurately shifts \( \boldsymbol{p}^{i}_k \) has only radial and tangential components, \( v^{i}_k \) and \( u^{i}_k \), respectively, so that:
\begin{equation}\label{eq:accurate_shift}
\boldsymbol{q}^{i}_k = \boldsymbol{p}^{i}_k + v^{i}_k \hat{\boldsymbol{v}}^i_k (T_0 - T_k) + u^{i}_k \hat{\boldsymbol{u}}^i_k (T_0 - T_k),
\end{equation}
where \( \hat{\boldsymbol{v}}^i_k \) and \( \hat{\boldsymbol{u}}^i_k \) are 3D unit vectors in the radial and tangential directions on a plane (zero in height), as indicated by dashed arrows in \cref{fig:agg_method}(b) (with further explanation in \cref{sec:dopp_background}).
In \cref{eq:accurate_shift}, the radial velocity \( v^{i}_k \) is known from the dynamic Doppler measurement, while the tangential velocity \( u^{i}_k \) is unknown. Therefore, we approximate the shifted position \( \boldsymbol{q}^{i}_k \) by only compensating for the radial shift. This approximation, \( \tilde{\boldsymbol{q}}^{i}_k \), is given by:
\begin{equation}\label{eq:approx_shift}
\boldsymbol{q}^{i}_k\approx\tilde{\boldsymbol{q}}^{i}_k = \boldsymbol{p}^{i}_k + v^{i}_k \hat{\boldsymbol{v}}^i_k (T_0 - T_k).
\end{equation}
The shifted points, $\tilde{\boldsymbol{q}}^i_{-K}$ and $\tilde{\boldsymbol{q}}^j_{-1}$, from times $T_{-K}$ and $T_{-1}$, are shown in \cref{fig:agg_method}(b), with green and purple hollow circles, respectively. 
The distance offset between the accurate and approximated positions is given by:
\begin{equation}\label{eq:shift_err}
\epsilon^{i}_k = \| \boldsymbol{q}^{i}_k - \tilde{\boldsymbol{q}}^{i}_k \| = |u^i_k| (T_0 - T_k).
\end{equation}
By substituting \cref{eq:rad_tan_components} into \cref{eq:shift_err} we have that 
\begin{equation}\label{eq:shift_err_dopp}
\epsilon^{i}_k = |v^i_k| (T_0 - T_k)\tan(\theta^i_k + \alpha^i_k),
\end{equation}
where $\theta^i_k$, and $\alpha^i_k$, are the $i^{th}$ reflection point angle and heading direction at frame time $T_k$.
The offsets $\epsilon^i_{-K}$ and $\epsilon^i_{-1}$ for points $\tilde{\boldsymbol{q}}^{i}_{-K}$ and $\tilde{\boldsymbol{q}}^{i}_{-1}$ are indicated by dotted lines in \cref{fig:agg_method}(b) with purple and green colors, respectively.

The average offset is obtained by taking the expectation of \cref{eq:shift_err_dopp} with respect to $\alpha^i_k$. We set a pre-defined limit to the tolerated average offset, denoted by $\mathcal{D}$, such that
\begin{equation}\label{eq:err_exp_limit}
\mathbb{E}_{\alpha^i_k}\left[\epsilon^{i}_k\right] = |v^i_k| (T_0 - T_k) g(\theta^i_k)\leq \mathcal{D}
\end{equation}
where 
\begin{equation}\label{eq:tan_expectation}
g(\theta^i_k)=\mathbb{E}_{\alpha^i_k}\left[\tan(\theta^i_k + \alpha^i_k)\right]  \hspace{-0.1cm}= \hspace{-0.1cm} \int_{-\pi/2}^{\pi/2} \hspace{-0.3cm} \tan(\theta^i_k + \alpha) f_{\alpha^i_k}(\alpha) d\alpha
\end{equation}
and \( f_{\alpha^i_k}(\alpha) \) is the distribution of the dynamic vehicle orientation angle \( \alpha \) (shown in \cref{fig:dopp_components}), which can be empirically derived from an annotated dataset or approximated by a Laplace distribution. For further details on calculating of $g(\theta^i_k)$, please refer to \cref{sec:yaw_func}.

By rearranging terms in \cref{eq:err_exp_limit} we can express a limit on the aggregation duration per each point, depending on its Doppler and angle, that will ensure the average shift offset will be below the predefined tolerance $\mathcal{D}$ as follows:
\begin{equation}\label{eq:duration_limit}
T_0 - T_k \leq \frac{\mathcal{D}}{|v^i_k| g(\theta^i_k)}.
\end{equation}
In aggregating points from previous frames to the current time, we include only those with an aggregation duration ($T_0-T_k$) that satisfy \cref{eq:duration_limit}, ensuring the average shift offset remains within the specified tolerance $\mathcal{D}$. Note that as a result, each point has a specific aggregation duration, which differs from conventional methods that use a fixed aggregation duration to all points.

In the illustrated example of \cref{fig:agg_method}(b), the aggregation duration \( T_0 - T_{-1} \) results in an offset \( \epsilon^i_{-1} \) for \( \tilde{q}^i_{-1} \) that is below $\mathcal{D}$, so it is included in the aggregation at \( T_0 \). In contrast, \( \epsilon^i_{-K} \) for \( \tilde{q}^i_{-K} \) exceeds $\mathcal{D}$, hence this point is \emph{excluded}. The final aggregated result is shown in \cref{fig:agg_method}(c), this includes the current frame point and the previous frame point with low disparity. 
For comparison, \cref{fig:agg_method}(d) shows conventional aggregation without dynamic motion compensation, resulting in point spread along both radial and tangential directions. In contrast, our method increases the points density with minimal tangential spread and no radial spread. 

While \cref{fig:agg_method} illustrates a single reflection point from one dynamic object per time instance, the method applies to multiple reflection points from objects with varying velocities, including static ones. Each point is independently shifted along its radial direction to the ego-vehicle based on its dynamic Doppler measurement, with static points (zero dynamic Doppler) remaining unchanged. \cref{fig:qual_example} demonstrates \textit{DoppDrive}'s effectiveness in complex scenarios with multiple vehicles at different speeds.
The pseudocode for the proposed Doppler-Driven aggregation algorithm is provided in \cref{alg:dopp_based_agg}.
\begin{algorithm}
\caption{Doppler-Driven Aggregation (\textit{DoppDrive})}
\label{alg:dopp_based_agg}
\begin{algorithmic}[1]
\STATE \textbf{Setting:} Tolerated aggregation shift error, $\mathcal{D}$
\STATE \textbf{Input:} 
\begin{itemize}
    \item Set of points $\boldsymbol{p}^i_k$, with their dynamic Doppler velocity $v^i_k$, radial unit vector $\hat{\boldsymbol{v}}^i_k$, angle $\theta^i_k$, and frame time $T_k$
    \item Indices: $i \in \{0, 1, \dots, N\}$ (points), $k \in \{0, -1, \dots, -K\}$ (frame times)
\end{itemize}
\STATE \textbf{Output:} Aggregated set of shifted points $\mathcal{Q}$ across frame times $T_0, T_{-1}, \dots, T_{-K}$

\STATE \textbf{Initialize:} Aggregation set $\mathcal{Q} \gets \emptyset$

\FOR{$k = -K$ to $0$} 
    \FOR{$i = 0$ to $N$}
        \STATE Calculate $g(\theta^i_k)$ from \cref{eq:tan_expectation}
        \IF{$T_0 - T_k \leq \frac{\mathcal{D}}{|v^i_k| g(\theta^i_k)}$} 
            \STATE Shift point: $\tilde{\boldsymbol{q}}^{i}_k = \boldsymbol{p}^{i}_k + v^{i}_k \hat{\boldsymbol{v}}^i_k (T_0 - T_k)$
            \STATE Append $\tilde{\boldsymbol{q}}^{i}_k$ to aggregation set $\mathcal{Q}$
        \ELSE
            \STATE Discard point $\boldsymbol{p}^i_k$
        \ENDIF
    \ENDFOR
\ENDFOR

\RETURN Aggregated set of points $\mathcal{Q}$
\end{algorithmic}
\end{algorithm}

\subsection{Object Detection}
The $\textit{DoppDrive}$ point cloud aggregation, as  described in \cref{alg:dopp_based_agg}, operates frame by frame. For each new frame, a set of aggregated points, $\mathcal{Q}$, is computed in a sliding window fashion, combining points from the current frame and previous frames. For object detection, each point in $\mathcal{Q}$ is represented by six features: its 3D position $(x, y, z)$ in the radars coordinate system of the current frame time $T_0$, the dynamic Doppler component, reflection intensity, and the time frame index $k$ when the point was originally received. 
Any object detection network designed for point clouds can then be used to detect objects from the aggregated points of each frame, utilizing these input features.

\section{Long Range Radar Dataset - \textit{LRR-Sim}}\label{sec:LLR_SIM}
Our Doppler-Driven aggregation method addresses the increasing sparsity of radar point clouds at long ranges, therefore we are particularly interested in evaluating its performance at long  distances. Existing radar point cloud datasets cover short and medium range distances, reaching up to $175m$ \cite{matuszka2022aimotive}. However, there is a gap in datasets for long-range automotive radar \cite{continental_ars540,bosch_front_radar,zf_4d_radar_2021,aptiv_radars}, which can detect up to $300m$, a range critical for autonomous driving on highways and at high speeds \cite{sun2020mimo,markel2022radar,zhou2020mmw}. Obtaining accurate annotations beyond $175m$ is challenging, as the radar’s point cloud becomes very sparse at these distances, and typical LIDAR and camera sensors used for annotations lack this range. 
To address this, we introduce a long-range radar dataset with point clouds up to $300m$ and accurate annotations generated via radar simulation, building on prior work that demonstrates the realism and reliability of simulated radar data \cite{bialer2024radsimreal}.

The simulation models a high-end, long-range radar in highway scenarios. Automotive scenarios were generated in CARLA \cite{dosovitskiy2017carla} using CARLA highway environment maps, and converted to radar point clouds using a physical model of a 77GHz MIMO radar with 12 transmit and 16 receive antennas. The radar’s received signal was computed via ray tracing between the radar antennas and objects in the scene, followed by range-Doppler 2D FFT and azimuth-elevation beamforming to obtain the reflection intensity spectrum. Point clouds were then extracted from this spectrum using the CFAR algorithm \cite{rohling1983radar}. Additional implementation details are provided in \cref{sec:lrrSim_supp}.

We refer to the long-range radar simulation dataset as \textit{LRR-Sim}. It includes 42 training and 8 testing highway scenarios, each lasting approximately 30 seconds at 20 frames per second, resulting in 18,172 training frames and 3,363 testing frames. Vehicle types include cars, trucks, and vans, with an average of 7.3 vehicles per frame and a total of $\sim 160K$ vehicle instances with $\sim 28K$ beyond $175m$. 
The point cloud extends up to $300m$ with a field of view of $\pm 55^{\circ}$ in azimuth, and $\pm20^{\circ}$ in elevation. All vehicles within this range and field of view are accurately annotated with 3D bounding boxes.  
For more details on \textit{LRR-Sim} and example demonstrations, please refer to our GitHub\footnote{The \textit{LRR-Sim} dataset is available at: \url{https://github.com/yuvalHG/LRRSim}}. We hope this dataset serves as a valuable resource for advancing radar research, including applications beyond the scope of this paper.

%% file: 4_results.tex
\section{Results}
\label{sec:results}
We evaluate the performance improvement of object detection when using \textit{DoppDrive} as a preprocessing step with four radar-based detection networks across three datasets, covering diverse conditions and settings.
The radar detection networks include Radar PillarNet (RPNet) \cite{zheng2023rcfusion}, based on PointPillars \cite{lang2019pointpillars}; SMURF (SMF) \cite{liu2023spatial}, which enhances PointPillars with kernel density estimation; Nvradarnet (NVR) \cite{popov2023nvradarnet}, a 2D convolutional Feature Pyramid Network in bird’s-eye view; and K-Radar (KRD) \cite{paek2023enhanced}, which utilizes 3D sparse convolution. Implementation details are provided in \cref{sec:detctor_det_supp}.

We selected datasets with large detection ranges to highlight the role of aggregation in addressing long-range sparsity, and with varying point cloud densities due to differences in radar resolution. \cref{tab:dataset_compare} compares the datasets used in our evaluation. aiMotive has the lower angular resolution and sparsest point cloud compared to RADIal and \textit{LRR-Sim}; all are sparser than LiDAR. Detection ranges reach up to $175m$ for aiMotive, $103m$ for RADIal, and $300m$ for \textit{LRR-Sim}, meeting the requirements for long-range automotive radar \cite{continental_ars540,bosch_front_radar,zf_4d_radar_2021,aptiv_radars,sun2020mimo,markel2022radar,zhou2020mmw}. Velocity sources also differ: aiMotive uses INS-GPS, RADIal lacks accurate INS and uses radar-based estimation, and \textit{LRR-Sim} provides ground-truth from simulation. The datasets span diverse scenarios: RADIal includes city, countryside, and highway scenes; aiMotive covers urban and highway; and \textit{LRR-Sim} focuses on highways.

\begin{table*}[t]
\centering
\setlength{\tabcolsep}{3pt} 
\begin{tabular}{c|cccc|cccc|cccc}
\toprule
\multirow{2}{*}{Method} & \multicolumn{4}{c|}{aiMotive} & \multicolumn{4}{c|}{Radial} & \multicolumn{4}{c}{LRR-Sim} \\
 & SMF & KRD & RPNet & NVR & SMF & KRD & RPNet & NVR & SMF & KRD & RPNet & NVR \\
\hline\hline
No Agg. & 74.6 & 75.0 & 73.4 & 70.9 & 91.3 & 91.4 & 91.0 & 89.3 & 89.2 & 88.7 & 88.7 & 86.6 \\
Standard Agg. & 81.7 & 82.3 & 80.2 & 75.2 & 92.5 & 92.1 & 92.3 & 91.2 & 91.1 & 89.9 & 90.8 & 90.1 \\
\textit{DoppDrive} & \textbf{89.1} & \textbf{89.4} & \textbf{87.8} & \textbf{81.8} & \textbf{95.8} & \textbf{95.5} & \textbf{95.7} & \textbf{94.1} & \textbf{93.3} & \textbf{93.1} & \textbf{92.9} & \textbf{92.7} \\
\bottomrule
\end{tabular}
\vspace{-8pt}
\caption{Average Precision (AP) of four detectors: SMURF (SMF) \cite{liu2023spatial}, K-Radar (KRD) \cite{paek2023enhanced}, Radar PillarNet (RPNet) \cite{zheng2023rcfusion}, and Nvradarnet (NVR) \cite{popov2023nvradarnet}. Results are evaluated across three datasets (aiMotive, Radial, LRR-Sim) with no aggregation, standard aggregation with ego-vehicle motion compensation, and \textit{DoppDrive}.}
\vspace{-5pt}
\label{tab:results}
\end{table*}

In all tests, we set the tolerance disparity distance to $\mathcal{D}=2m$ to limit the aggregation duration in \textit{DoppDrive}. This value was empirically found to give the best results, as detailed 
in \cref{sec:optimize_D}. We compare \textit{DoppDrive} with two reference methods: (1) standard aggregation, which combines points from frames up to $0.7s$ in the past with ego-vehicle motion compensation, an empirically optimized duration for best performance; and (2) no aggregation. In both references, unless specified otherwise, the ego-speed Doppler component is removed from the point cloud Doppler measurements. This aids object detection by distinguishing reflections from static objects (near-zero Doppler) and dynamic objects (non-near-zero Doppler). 

We evaluated object detection performance using average precision (AP), the integral of the precision-recall curve, for the 'vehicle' class, which includes cars, trucks, vans, and buses. Detection accuracy is measured by 2D bounding box IOU in birds-eye-view between detections and ground truth, using a 0.1 threshold. 
While aiMotive and \textit{LRR-Sim} provide bounding box annotations, RADIal offers only a single point on the object's front-facing contour. To enable consistent evaluation, we assigned a fixed-size bounding box—based on average vehicle dimensions—so that its front edge aligns with the provided point, assuming zero heading angle.

\begin{table}[h]
\centering
\begin{tabular}{l|c|c|c}
\toprule
 & aiMotive & RADIal & LRR-Sim \\
\hline \hline
Resolution  & Low & High & High \\

Range & $175m$ & $103m$ & $300m$ \\

FOV & $\pm 16^{\circ}$ & $\pm 90^{\circ}$ & $\pm 55^{\circ}$ \\

Ego-vel & INS-GPS & Radar & Simulation \\

FPS & $18$ & $5$ & $20$ \\

\# Train & $21402$ & $5450$ & $18172$ \\

\# Test & 5181 & 1873 &3363 \\
\# 3D Boxes & $427K$ & $9.5K$ & $160K$ \\
\bottomrule
\end{tabular}
\vspace{-5pt}
\caption{Comparison of the three tested datasets by angular resolution (Resolution), point cloud maximal range (Range), azimuth field-of-view (FOV), ego-vehicle velocity source (Ego-vel), frame rate (FPS), the number of frames in the training (\# Train) and test (\# Test) sets, and number of objects (\# 3D Boxes).}
\vspace{-5pt}
\label{tab:dataset_compare}
\end{table}

\subsection{Object Detection Performance Gain}
\cref{tab:results} compares the performance of the four selected object detection methods across three datasets, all detailed in \cref{sec:results}. The evaluation uses three types of point cloud inputs: (1) single-frame reflection points without aggregation, (2) standard aggregation with ego-motion compensation, and (3) our Doppler-driven aggregation method, \textit{DoppDrive}.
For all methods, ego-speed Doppler components were removed from the Doppler measurements, and the input features of each point included the $x,y,z$ coordinates, Doppler, reflection intensity, and time index for aggregated inputs. The results show that both aggregation methods outperform single-frame detection, with Doppler-driven aggregation achieving a greater improvements over standard aggregation across all detectors and datasets. 
The largest performance gain is observed on the aiMotive dataset, likely due to its radar’s lower angular resolution and sparser point cloud compared to the other datasets. Because the point cloud is more sparse, it benefits more from \textit{DoppDrive}'s aggregation, which increases its density with minimal scatter. Notably, the performance improvement is evident not only when precise ego-vehicle velocity is provided by GPS-INS (aiMotive) or simulation (\textit{LRR-Sim}) but also when the velocity is estimated directly from the radar point cloud (RADIal). Among the four detectors compared, SMF \cite{liu2023spatial} and KRD \cite{paek2023enhanced} achieve the highest performance, both with and without aggregation. Qualitative examples demonstrating \textit{DoppDrive}'s effectiveness appear in \cref{fig:qual_example} and in \cref{sec:addtional_res_supp}.  

To analyze the performance gains, we present an ablation study in \cref{tab:ablations} using the aiMotive dataset and SMF \cite{liu2023spatial} detector, assessing the contributions of each component in \textit{DoppDrive} as outlined in the table columns: (1) Ego-vehicle velocity compensation in aggregation (Ego. agg.), (2) GPS-INS ego-vehicle velocity measurement or radar-based estimation (GPS-INS Vel.), (3) Ego-speed Doppler component removal (Ego-Dopp.), (4) radial motion compensation using \cref{eq:approx_shift} (Radial Comp.), and (5) tangential spread mitigation by limiting aggregation duration based on \cref{eq:duration_limit} (Non-fix Duration).

The ablation test shows that both radial compensation and variable aggregation durations per point contribute to \textit{DoppDrive}'s performance gains (rows 6 vs. 5, and 8 vs. 6, respectively). 
It also shows that using the ego-vehicle velocity estimated from the radar point cloud, instead of the GPS-INS velocity measurement, results in a relatively small performance degradation for \textit{DoppDrive} (row 7 vs. 8). 
\subsection{Long Range Detection Performance}
In this section, we evaluate \textit{DoppDrive} at long ranges ($175m$-$300m$), where the point cloud is highly sparse. Detecting objects at these distances is crucial for high-speed autonomous driving. Since datasets like aiMotive and RADIal are limited to $175m$, we use \textit{LRR-Sim} (\cref{sec:LLR_SIM}), which provides simulated long-range radar data with accurate annotations up to $300m$.

\cref{tab:LRR_sim_res} shows detection performance on \textit{LRR-Sim} for objects within $175m$-$300m$ using four object detectors from \cref{sec:results}. We compare three input types: \textit{DoppDrive}, standard aggregation with ego-motion compensation, and no aggregation, with the ego-speed Doppler component removed in all cases. Compared to the full-range results in \cref{tab:results}, accuracy drops significantly at longer distances, but \textit{DoppDrive} retains and even enhances its advantage. This improvement at longer ranges is due to the sparser point cloud, where increased density from temporal aggregation with minimal scatter proves even more beneficial.

\begin{table}[h]
\centering
\renewcommand{\arraystretch}{1.2} 
\begin{tabular}{>{\centering\arraybackslash}p{0.25cm}>{\centering\arraybackslash}p{0.5cm}>{\centering\arraybackslash}p{1.2cm}>{\centering\arraybackslash}p{0.75cm}>{\centering\arraybackslash}p{1cm}>{\centering\arraybackslash}p{1.2cm}|>{\centering\arraybackslash}p{0.5cm}}
\toprule
 ID &\small \centering Ego. agg. & \small GPS-INS Vel. & \small Ego-Dopp. & \small Radial Comp. & \small Non-fix Duration & \small AP \\
\hline\hline
1 & \X & - & \X & \X & \X & 66.3 \\
2 & \X & \X & \V & \X & \X & 73.1 \\
3 &\X & \V & \V & \X & \X & 74.8 \\
4 &\V & \V & \X & \X & \X & 76.1  \\
5 &\V & \V & \V & \X & \X & 81.4  \\
6 &\V & \V & \V & \V & \X & 85.2  \\
7 &\V & \X & \V & \V & \V & 87.7  \\
8 &\V & \V & \V & \V & \V & \textbf{89.1}  \\
\bottomrule
\end{tabular}
\vspace{-5pt}
\caption{Ablation study of \textit{DoppDrive} components: Average Precision (AP) on aiMotive dataset with SMF detector \cite{liu2023spatial}.}
\vspace{-5pt}
\label{tab:ablations}
\end{table}

\begin{table}[h]
\centering
\begin{tabular}{c|cccc}
\toprule
 Method & SMF & KRD & RPNet & NVR \\
\hline\hline
No Agg. & 61.2 & 61.5 & 61.8 & 60.7 \\
Standard Agg.        & 63.1 & 64.1 & 62.7 & 61.3 \\
\textit{DoppDrive} & \textbf{69.2} & \textbf{68.9} & \textbf{68.6} & \textbf{66.9} \\
\bottomrule
\end{tabular}
\vspace{-5pt}
\caption{Average Precision at range $> 175m$ on the \textit{LRR-Sim} dataset with four detectors: SMF \cite{liu2023spatial}, K-Radar (KRD) \cite{paek2023enhanced}, RPNet \cite{zheng2023rcfusion}, and Nvradarnet (NVR) \cite{popov2023nvradarnet}.}
\vspace{-10pt}
\label{tab:LRR_sim_res}
\end{table}

\subsection{Optimization of the Dispersion Tolerance \texorpdfstring{$\mathcal{D}$}{D}}\label{sec:optimize_D}
We perform an ablation study on \textit{DoppDrive}'s hyperparameter $\mathcal{D}$, which defines the tolerated average position offset during aggregation and controls the aggregation duration for each point. \cref{tab:d_ablation} shows the average precision performance of the SMF \cite{liu2023spatial} object detection network with the \textit{LRR-Sim} dataset for different values of \(\mathcal{D}\). The results indicate that the best performance was achieved with $\mathcal{D}=2m$, which was used for all tests presented in the paper.
\begin{table}[!h]
\begin{center}
\setlength{\tabcolsep}{0.8mm}
\begin{tabular}{l|ccccc}
\toprule
$\mathcal{D}$ [m]& 1 & 2 & 3 & 4 & 5 \\
\midrule
AP & 91.2 & \textbf{93.3} & 92.5 & 92.3 & 89.6 \\
\bottomrule
\end{tabular}
\end{center}
\vspace{-6mm}
\caption{Average precision (AP) of SMF \cite{liu2023spatial} on the \textit{LRR-Sim} dataset as a function of the aggregation displacement tolerance $\mathcal{D}$.}
\vspace{-4mm}
\label{tab:d_ablation}
\end{table}
\begin{figure}
    \vspace{+2pt}
    \centering
    \includegraphics[width=0.9\linewidth]{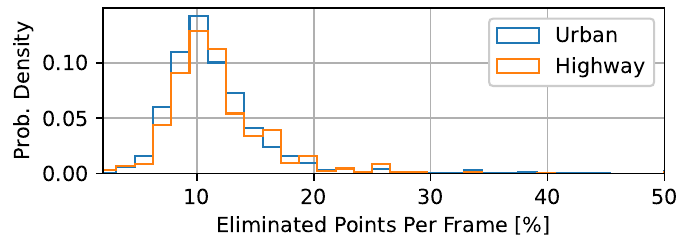}
    \vspace{-12pt}
    \caption{Point elimination in urban and highway (aiMotive)}
    \label{fig:elim_hist}
    \vspace{-15pt}
\end{figure}
\subsection{Point Elimination Evaluation}
\textit{DoppDrive} shortens integration duration to filter out points with high tangential dispersion, resulting in point elimination compared to fixed-duration aggregation. 
\cref{fig:elim_hist} shows that \textit{DoppDrive} removes an average of 11\% of dynamic points per frame, relative to fixed-duration aggregation, in aiMotive’s urban and highway scenes, with most frames exhibiting less than 20\% elimination.
This yields a 3.9-point AP gain (rows 8 vs. 6 in \cref{tab:ablations}), demonstrating that even modest elimination improves performance. A detailed breakdown of point elimination is provided in \cref{sec:point_elm_supp}.

%% file: 5_conclusion.tex
\section{Conclusion}\label{sec:conclusion}
In this paper, we present \textit{DoppDrive}, a novel method to enhance radar point cloud density before object detection with minimal scatter by leveraging Doppler information in temporal aggregation. \textit{DoppDrive} reduces scatter from dynamic objects by adjusting previous frames’ points in the radial direction based on their dynamic Doppler components and by limiting each point's aggregation duration according to its specific parameters. An ablation study validated the contribution of both elements, while complexity analysis showed minimal computational overhead. \textit{DoppDrive}'s enhanced point cloud improves object detection performance, as demonstrated across four detectors and three datasets.

Additionally, we introduce \textit{LRR-Sim}, a long-range radar simulation dataset with sparse point clouds resulting from extended detection ranges, and precise annotations up to $300m$, significantly beyond existing datasets. The dataset is publicly available to support advancements in long-range radar detection, which is essential for high-speed autonomous driving.

%% file: 6_supp.tex
\input{sup_sec_qualitative}
\begin{figure*}[h]
\centering
\includegraphics[width=1\textwidth]{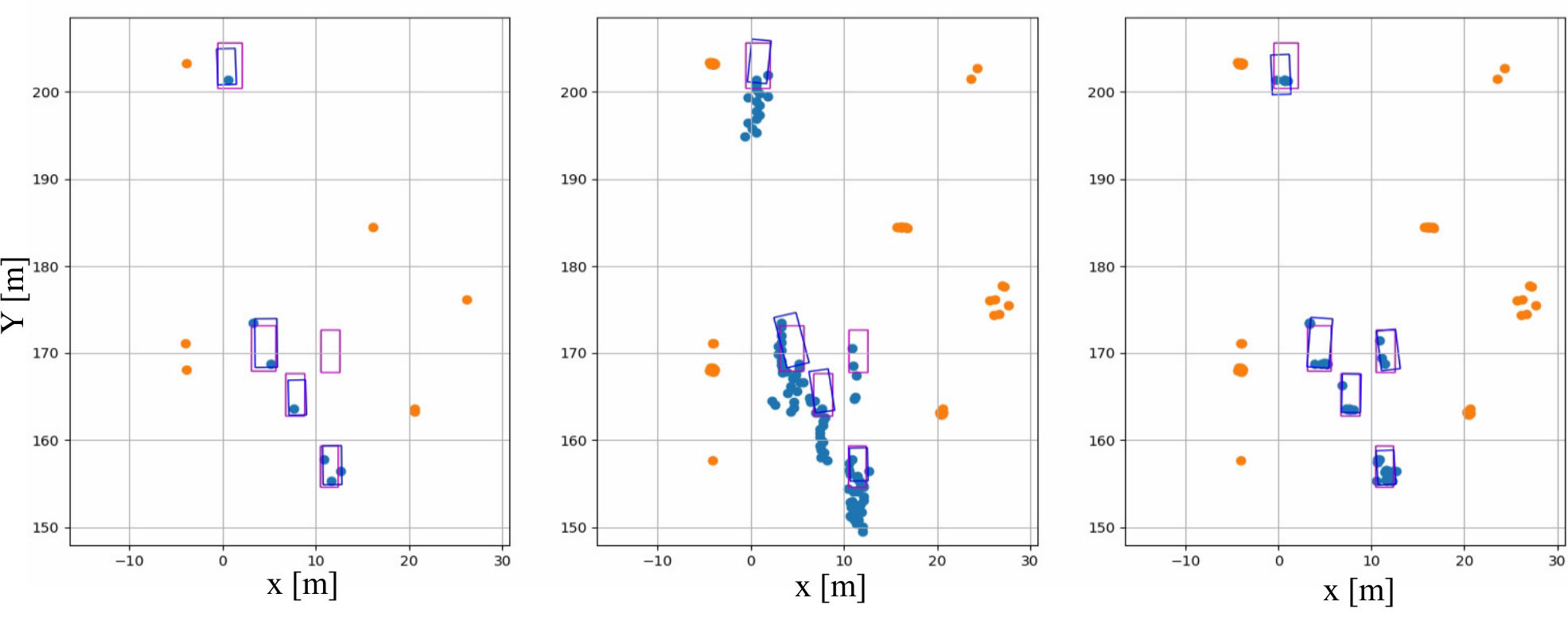}
\caption{Qualitative example from \textit{LRR-Sim} shown in bird’s-eye view. The point cloud of dynamic objects is shown in blue, while that of static objects is shown in orange. SMF \cite{liu2023spatial} detections are represented by blue bounding boxes, and ground truth bounding boxes are shown in pink. The left figure depicts results without aggregation, the middle figure shows standard aggregation with ego-motion compensation, and the right figure shows the results of \textit{DoppDrive}.}
\label{fig:carla_qualitative1}
\end{figure*}
\begin{figure*}[h]
\centering
\includegraphics[width=1.0\textwidth]{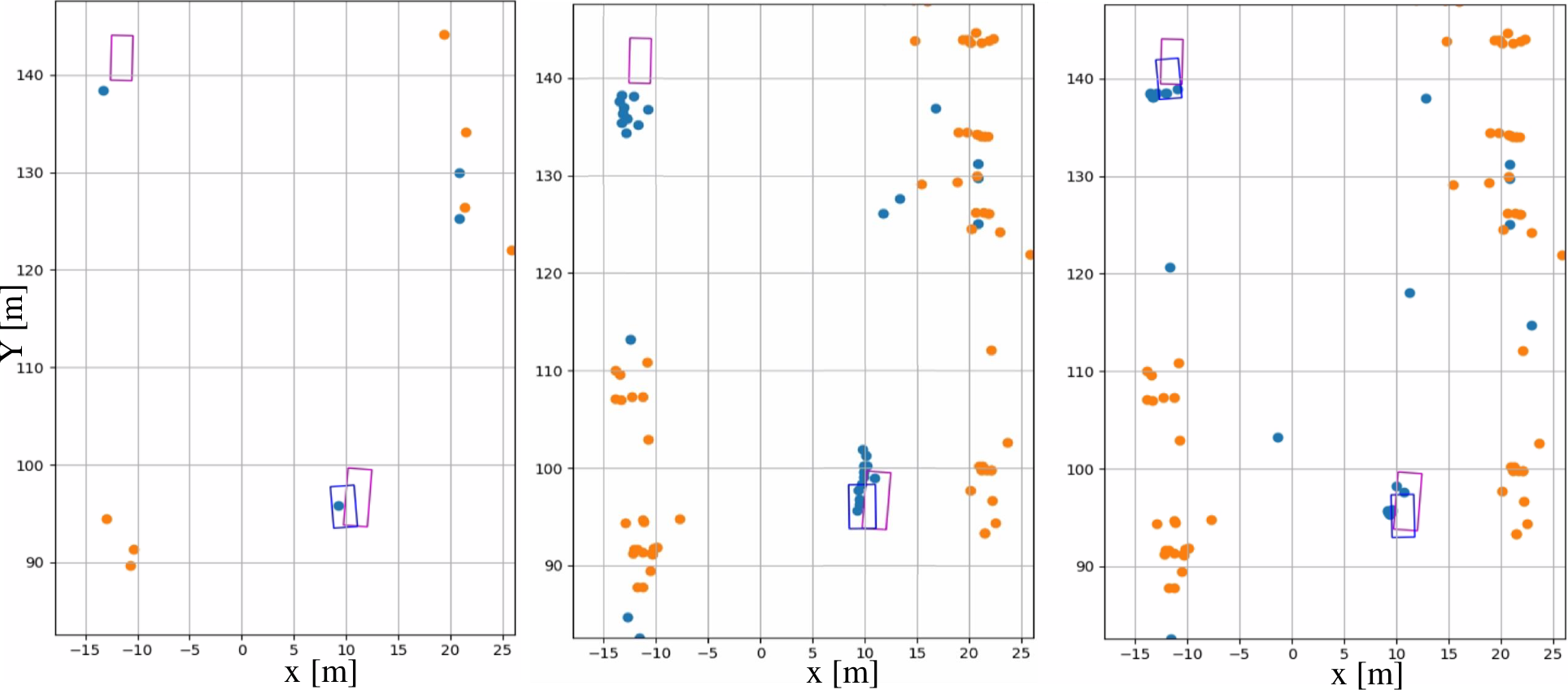}
\caption{Qualitative example from aiMotive shown in bird’s-eye view. The point cloud of dynamic objects is shown in blue, while that of static objects is shown in orange. SMF \cite{liu2023spatial} detections are represented by blue bounding boxes, and ground truth bounding boxes are shown in pink. The left figure depicts results without aggregation, the middle figure shows standard aggregation with ego-motion compensation, and the right figure shows the results of \textit{DoppDrive}.}
\label{fig:aiMotive_qualitative1}
\end{figure*}
\begin{figure*}[h]
\centering
\includegraphics[width=1.0\textwidth]{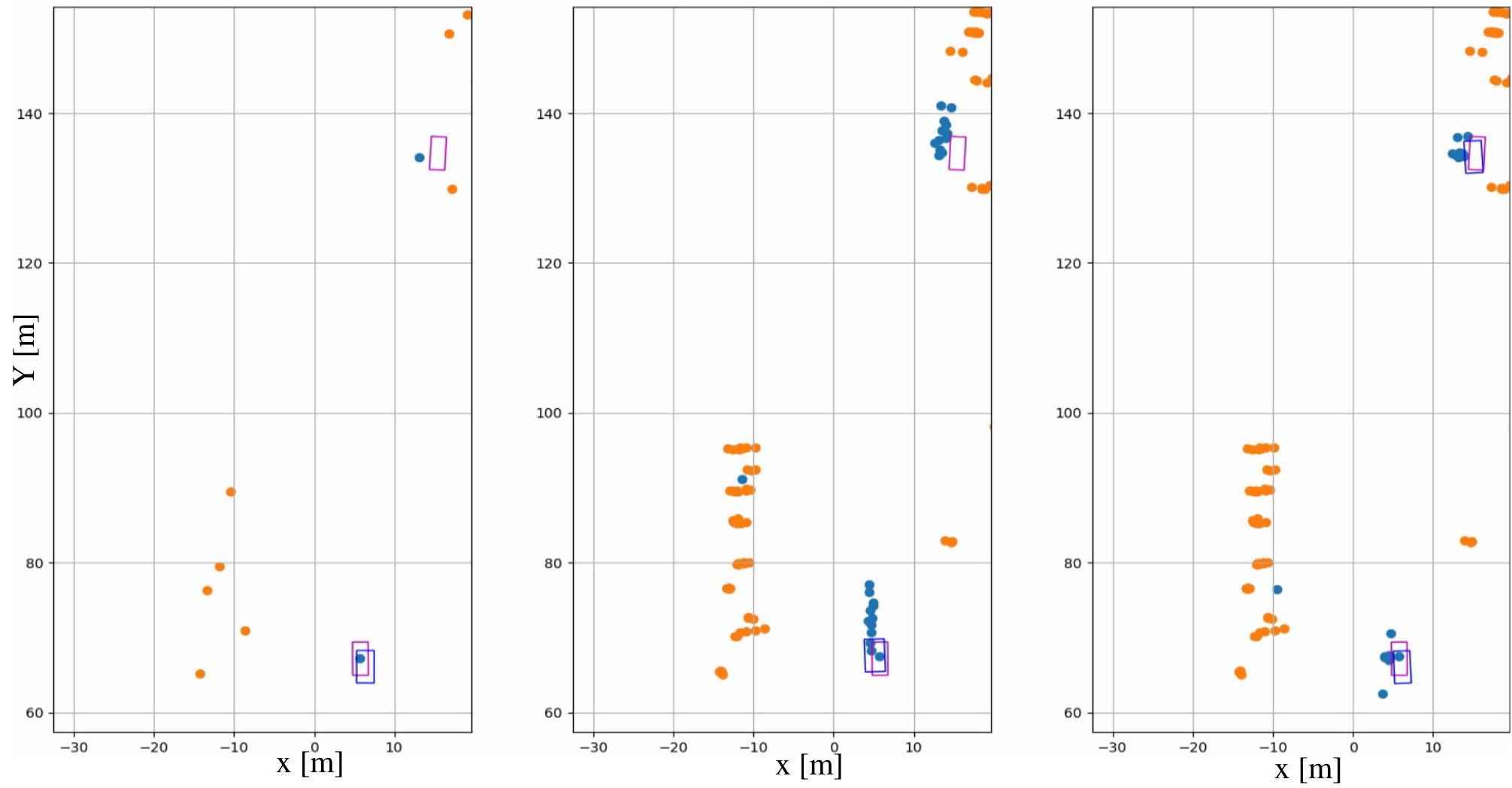}
\caption{Qualitative example from aiMotive shown in bird’s-eye view. The point cloud of dynamic objects is shown in blue, while that of static objects is shown in orange. SMF \cite{liu2023spatial} detections are represented by blue bounding boxes, and ground truth bounding boxes are shown in pink. The left figure depicts results without aggregation, the middle figure shows standard aggregation with ego-motion compensation, and the right figure shows the results of \textit{DoppDrive}.}
\label{fig:aiMOtive_qualitative2}
\end{figure*}

\input{sup_sec_computation_complexity}
\input{sup_sec_impact_on_point_elimination}
\input{sup_sec_yaw_dist}
\input{sup_sec_lrr_sim}
\input{sup_sec_detectors_details}

\input{sup_sec_vertical_velocity}

%% file: sup_sec_qualitative.tex
\section{Additional Qualitative Results}\label{sec:addtional_res_supp}

In this section, we provide qualitative examples supplementary to those shown in \cref{fig:qual_example}. \cref{fig:carla_qualitative1}  showcase an example from \textit{LRR-Sim} dataset, while \cref{fig:aiMotive_qualitative1,fig:aiMOtive_qualitative2} present two examples from the aiMotive dataset. The layout follows the same structure as \cref{fig:qual_example}.
For each example, the single-frame point cloud (without aggregation) is shown on the left, standard aggregation over 0.7s is displayed in the middle, and the results of \textit{DoppDrive} are presented on the right. Reflection points from dynamic and static objects are plotted in blue and orange, respectively. Detected bounding boxes using SMF \cite{liu2023spatial} are marked in blue, while ground truth bounding boxes are marked in pink. 

In all the examples, it is evident that without aggregation, the reflection points are sparse, often with only one or two points per object, and sometimes none, leading to missed object detections. Standard aggregation results in significant scatter of reflection points from dynamic objects (blue points). In contrast, \textit{DoppDrive} produces dense reflection points with minimal scatter, resulting in detections that are better aligned with the annotations compared to standard aggregation.

%% file: sup_sec_computation_complexity.tex
\section{Computational Complexity Evaluation}
We next assess the computational overhead of \textit{DoppDrive}, implemented according to \cref{alg:dopp_based_agg}, with $g(\theta)$ precomputed for each $\theta$ and accessed via a lookup table. \cref{tab:compute_table} compares the runtime of SMF \cite{liu2023spatial} object detection with \textit{DoppDrive} aggregation, standard aggregation, and no aggregation on an NVIDIA Tesla V100 SXM2 32 GB. The results show that \textit{DoppDrive}'s aggregation time is significantly smaller than the detection time, contributing minimal runtime overhead and making it suitable for real-time applications. 
\textit{DoppDrive} slightly reduces detector runtime compared to standard aggregation by aggregating fewer points due to its dynamic aggregation duration.
\begin{table}[h]
\centering
\renewcommand{\arraystretch}{1.0} 
\setlength{\tabcolsep}{4pt} 
\begin{tabular}{c|c|c|c}
\hline
\textbf{Aggregation} & \textbf{Aggregation} & \textbf{Detection} & \textbf{Total}  \\
\textbf{Method}  & \textbf{Time [ms]} & \textbf{Time [ms]} & \textbf{Time [ms]} \\ \hline
None &  0 & 66.4 & 66.4 \\
Standard &  1.4 & 67.8 & 69.2 \\
\textit{DoppDrive}  &  2.0 & 67.5 & 69.5 \\\hline
\end{tabular}
\caption{Runtime assessment of \textit{DoppDrive} using the SMF detector \cite{liu2023spatial}. The columns, from left to right, show the point cloud aggregation methods, aggregation runtime, object detection runtime with the aggregated points, and the total runtime, calculated as the sum of aggregation and detection runtimes.}
\label{tab:compute_table}
\end{table}

%% file: sup_sec_impact_on_point_elimination.tex
\section{Impact of Point Elimination} \label{sec:point_elm_supp}
\textit{DoppDrive} limits integration duration to suppress points with high tangential dispersion, resulting in point elimination relative to fixed-duration aggregation. This subsection evaluates the elimination rate and its impact on performance.

\cref{tab:rebuttal_table} breaks down the AP gains and the average percentage of eliminated dynamic points per frame—relative to fixed-duration aggregation with radial compensation (rows 8 vs. 6 in \cref{tab:ablations})—across heading, velocity, and range bins. Elimination reaches up to 16\%, increasing at short ranges, higher speeds, and more tangential headings. While the overall elimination rate is modest, it has a noticeable impact on performance, with larger gains observed at higher elimination levels.
\begin{table}[t]
\renewcommand{\arraystretch}{1}
\setlength{\tabcolsep}{1.5pt}
\centering
\begin{tabular}{@{}lcccc@{}}
\toprule
Range $[m]$ & $[0,40]$ & $[40,80]$ & $[80,120]$ & $[120,160]$ \\
\textcolor{blue}{Elim.\%}/\textcolor{ForestGreen}{AP Gain} & \textcolor{blue}{16.6}/\textcolor{ForestGreen}{5.6} & 
\textcolor{blue}{12.4}/\textcolor{ForestGreen}{4.6} & 
\textcolor{blue}{10.9}/\textcolor{ForestGreen}{2.9} & 
\textcolor{blue}{10.5}/\textcolor{ForestGreen}{2.7} \\
\hline \hline
Abs. Velocity $[m/s]$ & $[0,12]$ & $[12,24]$ & $[24,36]$ & $[36,48]$ \\
\textcolor{blue}{Elim.\%}/\textcolor{ForestGreen}{AP Gain} & \textcolor{blue}{10.6}/\textcolor{ForestGreen}{3.7} & 
\textcolor{blue}{11.1}/\textcolor{ForestGreen}{4.3} & 
\textcolor{blue}{11.8}/\textcolor{ForestGreen}{4.9} & 
\textcolor{blue}{12.6}/\textcolor{ForestGreen}{5.3} \\
\hline \hline
Abs. Heading $[^\circ]$ & $[0,30]$ & $[30,60]$ & $[60,120]$ & $[120,180]$ \\
\textcolor{blue}{Elim.\%}/\textcolor{ForestGreen}{AP Gain} & \textcolor{blue}{10.0}/\textcolor{ForestGreen}{3.9} & 
\textcolor{blue}{11.0}/\textcolor{ForestGreen}{4.1} & 
\textcolor{blue}{15.5}/\textcolor{ForestGreen}{5.1} & 
\textcolor{blue}{11.0}/\textcolor{ForestGreen}{4.0} \\
\bottomrule
\end{tabular}
\vspace{-8pt}
\caption{Point elimination rate (\textcolor{blue}{blue}) and AP gain (\textcolor{ForestGreen}{green}) over radial compensation with fixed aggregation duration and no elimination, at different range, velocity, and heading intervals (aiMotive).}
\label{tab:rebuttal_table}
\vspace{-12pt}
\end{table}

%% file: sup_sec_yaw_dist.tex
\section{Calculation of $g(\theta^i_k)$}\label{sec:yaw_func}
\begin{figure}[b]
\centering
\includegraphics[width=1.0\columnwidth]{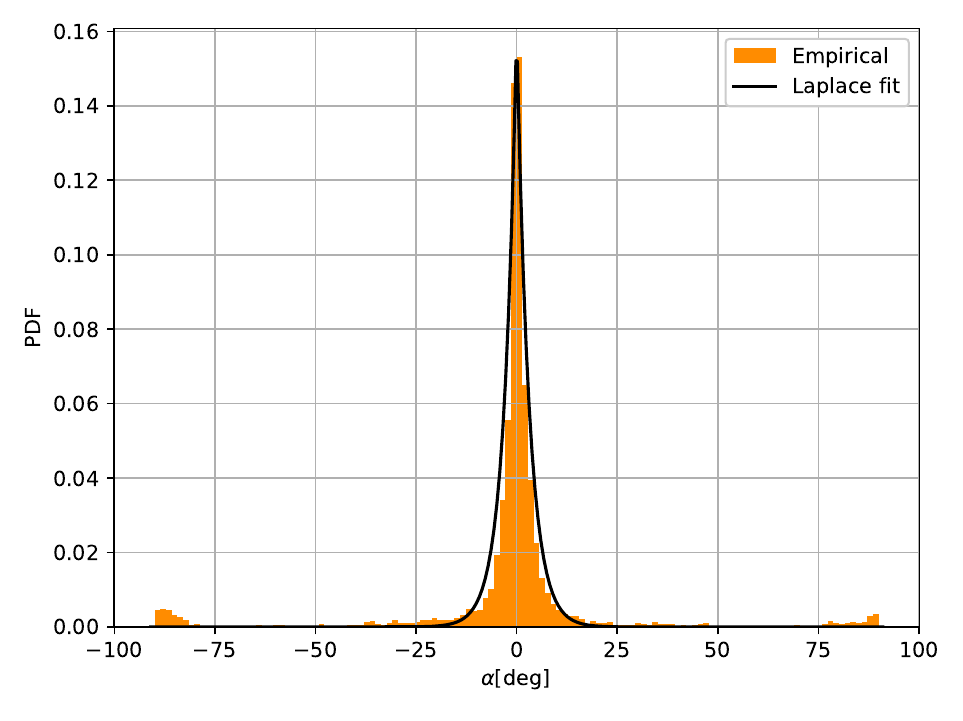}
\caption{Probability distribution function (PDF) of reflection points' heading directions used in calculating $g(\theta^i_k)$. The empirical PDF  from aiMotive dataset annotations is shown in orange, with the Laplace distribution fit plotted in black.}
\label{fig:aimotive_yaw}
\end{figure}

In \cref{eq:tan_expectation}, we provide the expression for $g(\theta^i_k)$, which is defined as: 
\begin{equation}\label{eq:g_eq}
g(\theta^i_k)=\mathbb{E}_{\alpha^i_k}\left[\tan(\theta^i_k + \alpha^i_k)\right]  \hspace{-0.1cm}= \hspace{-0.1cm} \int_{-\pi/2}^{\pi/2} \hspace{-0.3cm} \tan(\theta^i_k + \alpha) f_{\alpha^i_k}(\alpha) d\alpha,
\end{equation}
where $f_{\alpha^i_k}(\alpha)$ represents the probability distribution function of the heading angle $\alpha$. In this section, we explain how to compute $g(\theta^i_k)$ as defined by \cref{eq:g_eq}.

\cref{fig:aimotive_yaw} shows the probability density function (PDF) of \( \alpha \), derived from annotations in the aiMotive dataset. The distribution is sharply concentrated around zero, with significantly smaller secondary peaks near \( \pm 90^{\circ} \). To model this distribution, we fit a truncated Laplace distribution:
\begin{equation}\label{eq_laplace}
f_{\alpha^i_k}(\alpha) = \frac{1}{2b} \exp\left(-\frac{|\alpha - \mu|}{b}\right), \quad \alpha \in [-\pi/2, \pi/2],
\end{equation}
where we set \( \mu = 0 \) and \( b = 3.1 \). The minor peaks near \( \pm 90^{\circ} \) were neglected in this approximation. 
The fitted Laplace distribution is plotted in black in \cref{fig:aimotive_yaw}. As shown, it aligns well with the main mode of the empirical distribution. 

Based on this, the integral in \cref{eq:g_eq} can be evaluated numerically, either using the empirical histogram or the Laplace approximation for \( f_{\alpha^i_k}(\alpha) \). For the experiments in the main paper, we used the Laplace approximation to precompute \( g(\theta^i_k) \) over a grid of \( \theta^i_k \) values and stored the results in a look-up table for efficient use in \cref{alg:dopp_based_agg}. 

Using the full multi-modal histogram of \( \alpha \) from \cref{fig:aimotive_yaw} to compute the integral numerically yielded a slight performance improvement. Specifically, the SMF detector~\cite{liu2023spatial} achieved a 0.4-point increase in Average Precision on the aiMotive dataset.

%% file: sup_sec_lrr_sim.tex
\section{Supplementary Details on \textit{LRR-Sim}}\label{sec:lrrSim_supp}
\begin{figure}[h]
\centering
\includegraphics[width=1.0\columnwidth]{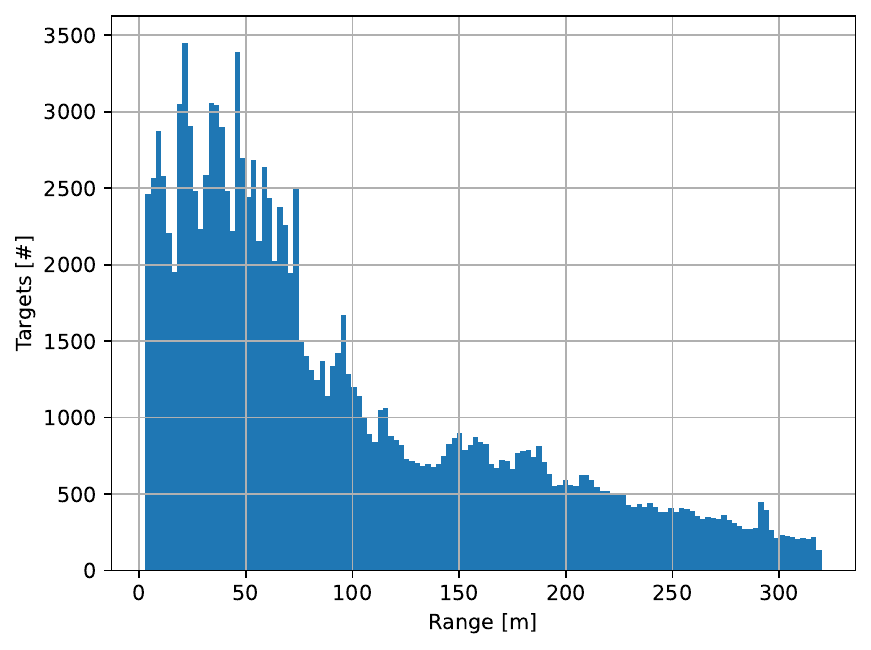}
\vspace{-20pt}
\caption{Number of vehicles by range in the \textit{LRR-Sim} dataset.}
\label{fig_lrr_sim_range}
\end{figure}

\begin{figure}[h]
\centering
\includegraphics[width=0.95\columnwidth]{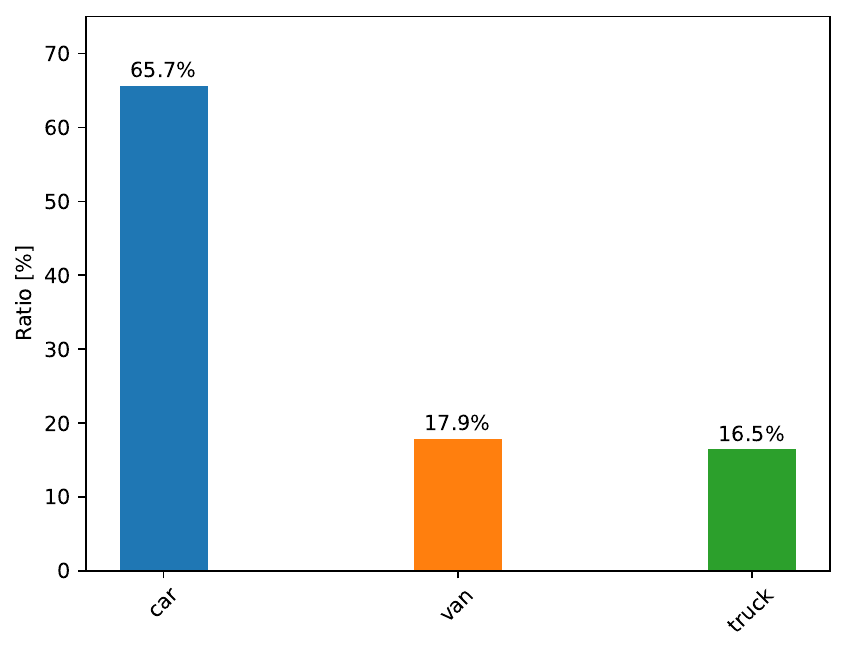}
\vspace{-15pt}
\caption{Distribution of vehicle types in the \textit{LRR-Sim} dataset.}
\label{fig_lrr_sim_class}
\end{figure}

In this section, we provide additional details about \textit{LRR-Sim}, the Long-Range Radar Simulation dataset introduced in \cref{sec:dopp_background}. 
\textit{LRR-Sim} focuses on highway scenarios with long-range vehicles of three types: 'car,' 'van,' and 'truck'. The number of vehicles as a function of range in the dataset is presented in \cref{fig_lrr_sim_range}, which indicating that vehicle ranges extend up to $300m$, with \(17.5\%\) of the vehicles located beyond $175m$. 
The distribution of vehicle types in the dataset is illustrated in \cref{fig_lrr_sim_class}. It is observed that \(65.7\%\) of the vehicles are of type 'car,' representing relatively small-sized vehicles, while \(17.9\%\) are medium-sized 'vans,' and \(16.5\%\) are large vehicles of type 'truck.'

The specifications of the simulated radar are detailed in \cref{tab_rad_spec}. These specifications align with those of high-end long-range automotive radars \cite{continental_ars540,zf_4d_radar_2021,aptiv_radars}, offering a detection range of up to $300m$ and high resolution in range, angle, and Doppler. 

The simulated radar features 12 transmit and 16 receive antennas and uses a fast chirp FMCW waveform. The highway environment and vehicles were modeled using the CARLA simulation platform \cite{dosovitskiy2017carla}, incorporating dense reflection points from objects. Ray tracing was performed to compute interactions between the antennas and reflection points in the environment.
The received signal at each antenna was the sum of the transmitted waveforms echoed by all reflection points in the scene, with adjustments for intensity and delay determined by each reflection point's intensity and distance. To obtain the radar reflection intensity spectrum in the range, Doppler, and angle dimensions, standard radar signal processing techniques were applied to the received signals. This included range FFT, Doppler FFT, and azimuth-elevation beamforming. Finally, the radar point cloud was generated by identifying spectrum peaks that exceeded a local noise threshold, using the CFAR algorithm \cite{rohling1983radar}. A video demonstration of \textit{LRR-Sim} is included our GitHub\footnote{The LRR-Sim dataset is available at: \url{https://github.com/yuvalHG/LRRSim}}.

\begin{table}[h]
  \centering
  \begin{tabular}{@{}lc@{}}
    \toprule
    Radar parameter &  \\
    \midrule
    Maximal range & $300m$ \\    
    Range resolution & $0.15m$ \\
    Azimuth field of view & $\pm55^{\circ}$\\
    Azimuth resolution & $1.2^{\circ}$ \\
    Elevation field of view & $\pm 20^{\circ}$\\
    Elevation resolution & $2^{\circ}$\\
    Doppler resolution & $0.13 m/s$ \\
    Doppler range & $[-80, 30] m/s$ \\
    \bottomrule
  \end{tabular}
  \caption{Specifications of radar used in \textit{LRR-Sim}}
  \label{tab_rad_spec}
\end{table}

\vspace{-10pt}

%% file: sup_sec_detectors_details.tex
\section{Detectors Implementation Details}\label{sec:detctor_det_supp}
For the performance evaluation of \textit{DoppDrive} in \cref{sec:results}, we used four detectors: Radar PillarNet (RPNet) \cite{zheng2023rcfusion}, SMURF (SMF) \cite{liu2023spatial}, Nvradarnet (NVR) \cite{popov2023nvradarnet}, and K-Radar (KRD) \cite{paek2023enhanced}. Details of their implementation are outlined below.
For all methods, ego-speed Doppler components were removed from the Doppler measurements, and the input features for each point included the $x,y,z$ coordinates, Doppler, reflection intensity, and time index for aggregated inputs. RPNet, SMF, and NVR used a $0.25m \times 0.25m$
input grid resolution in the X-Y plane, while KRD employed a $0.4m \times 0.4m \times 0.4m$ 3D grid resolution. Three anchor sizes were used to cover small, medium, and large vehicles, each with two orientations: $0^{\circ}$ and $90^{\circ}$. For SMF, we used two KDE blocks with bandwidths $R1=1m$ and $R2=2m$.
We apply radar point cloud augmentations, including flips, slight zoom in/out, and intensity jitter. Translations and rotations were excluded to prevent misalignment of Doppler measurements.

%% file: sup_sec_vertical_velocity.tex
\section{Vertical Velocity Extension}
In automotive scenarios, vertical displacement over short aggregation durations is typically negligible. Therefore, in deriving \textit{DoppDrive}, we assume zero vertical velocity and focus solely on longitudinal and lateral motion. However, \textit{DoppDrive} can be extended to account for vertical velocity if required. In this case, only the duration constraint in line 8 of Algorithm 1 is affected. The function $g$ would be updated to incorporate elevation angle and vertical heading distribution, mitigating dispersion in both vertical and tangential directions.